\documentclass[10pt,twocolumn,letterpaper]{article}

\usepackage[pagenumbers]{wacv} %

\usepackage{enumitem}
\usepackage{tikz}
\usetikzlibrary{positioning,fit,arrows.meta,shapes,backgrounds,calc}
\definecolor{wacvblue}{rgb}{0.21,0.49,0.74}
\usepackage[pagebackref,breaklinks,colorlinks,allcolors=wacvblue]{hyperref}
\usepackage{booktabs}
\usepackage{graphicx}
\def\wacvPaperID{*****} %
\def\confName{WACV}
\def\confYear{2027}

\title{WaveSpec: Multi-Spectral  Iris Presentation Attack Detection via Wavelength-Conditioned Spectral Coherence Adaptation}

\author{Raghavendra Ramachandra\\
SAFE Centre\\
Norwegian University of Science and Technology (NTNU), Norway\\
{\tt\small raghavendra.ramachandra@ntnu.no}
}

\begin{document}
\maketitle
\thispagestyle{empty}
{\let\thefootnote\relax\footnotetext{Accepted version of a paper to appear at the IEEE/CVF Winter Conference on Applications of Computer Vision (WACV) 2027. \copyright~2027 IEEE. Personal use of this material is permitted. Permission from IEEE must be obtained for all other uses, in any current or future media, including reprinting/republishing this material for advertising or promotional purposes, creating new collective works, for resale or redistribution to servers or lists, or reuse of any copyrighted component of this work in other works.}}

\begin{abstract}
Multispectral iris presentation attack detection benefits from observing
the same iris across multiple near-infrared wavelengths, but existing
methods often use wavelength information mainly for conditioning or
fusion. This can limit their ability to explicitly model whether the
spectral response of a sample is coherent across bands. In this paper,
we propose \textbf{WaveSpec}, a multispectral iris PAD framework that
performs feature-level spectral coherence adaptation. Each NIR band is
first encoded with a DINOv2 backbone. The proposed Set Conditioned
Spectral Coherence Adapter then updates each band feature using the
consensus of the available spectral set, the band-specific deviation
from that consensus, and wavelength identity. The adapted features are
projected into structural, spectral, and anomaly representations,
adaptively pooled, and classified as bona fide or attack. We evaluate
WaveSpec on the MSIrPAD benchmark under a strict cross-artefact protocol,
where the model is trained with one presentation attack instrument and
tested on unseen attacks. WaveSpec reduces the average D-EER from
8.45\% for SpectraIrisPAD to 3.33\%, and reduces average
BPCER@APCER=1\% from 43.16\% to 12.07\%. Ablation results show that the
SCSA adapter and spectral residual are central to the improvement, while
a band-subset analysis shows that WaveSpec remains effective under
partial spectral observations and performs best when all five bands are
available. The method adds only modest computational overhead, with
11.15 ms end-to-end latency per five-band sample on an NVIDIA L40S-48Q
GPU.
\end{abstract}

\section{Introduction}
\label{sec:intro}

Iris recognition is widely deployed because the iris offers rich,
stable, and highly distinctive texture, and most operational systems
acquire iris images in the near infrared range, where iris texture is
most visible. The same imaging process, however, exposes the system to
presentation attacks, where an attacker presents a printed iris, a
display image, or a textured contact lens to the sensor.

Iris presentation attack detection~\cite{IITDDB, Daugirisantispoof, NDDB, Gragnaniello2015, Raghavendra:2014:ESI:2683483.2683507, zhang2010contact, 7264974, kohli2013revisiting, komulainen2014generalized, yadav2018fusion, GLCM_IrisSpoof, DeepImage, wei2008counterfeit, raghavendra2017contlensnet, DLTrnasferLearning_CLDetection, agarwal2022generalizedIEEE,sharma2026towards, 11029274} has progressed from hand designed
texture and image quality features to deep neural networks and visual
foundation models. These methods can perform well when the training and
test conditions are similar, but their reliability often drops when the
test attacks differ from those seen during training. The LivDet Iris
evaluations have repeatedly shown that unseen presentation attack
instruments remain a major challenge~\cite{tinsley2023iris}. This is a
serious limitation for deployment because real attacks may differ in
material, printing process, lens pigment, display quality, and capture
condition.

A single near infrared image provides only one view of the eye and the
attack material, whereas multispectral imaging observes the same iris
across several wavelengths.  Bona fide irides vary across wavelength
according to tissue structure, pigmentation, and optical scattering,
while attack artefacts are governed by contact lens pigments, printing
substrates, inks, or display surfaces, often producing spectral behaviour
inconsistent with live tissue.  Prior multispectral iris PAD studies have
shown that NIR bands from roughly 800\,nm to 980\,nm improve the
detection of contact lens and print attacks~\cite{purkinje_Image,
MS_Iris_PAD,MS_Iris_PAD_contactLens,raghavendraMSIrisSensor,
ramachandra2024multi}.  SpectraIrisPAD introduced the MSIrPAD dataset
and a DINOv2-based multispectral PAD framework with spectral position
encoding and token fusion~\cite{ramachandra2025spectrairispad}; however,
it still leaves open the question of how to model whether the spectral
bands of the same sample agree or deviate before the final decision.

This paper proposes {WaveSpec}, a multispectral iris PAD model that
performs feature level spectral reasoning with a {Set Conditioned
Spectral Coherence Adapter} (SCSA). After DINOv2 extracts a feature from
each available wavelength, the SCSA updates each band feature using the
consensus of the spectral set, the deviation of that band from the
consensus, and a learnable wavelength identity, before pooling and
classification. This design allows the model to detect not only whether
a single wavelength looks suspicious, but also whether it behaves
consistently with the rest of the multispectral observation.

We evaluate WaveSpec on the MSIrPAD benchmark using a strict cross
artefact protocol, where the model is trained with one attack type and
tested on unseen attack types, comparing against six state-of-the-art
PAD baselines including SpectraIrisPAD
(Sec.~\ref{sec:exp_protocol}). The results show that explicitly
modelling spectral coherence improves generalisation to unseen iris
presentation attacks while adding only modest latency over the DINOv2
backbone.
\subsection{Contributions}
\label{sec:contributions}
The main contributions of this work are as follows:
\begin{enumerate} [leftmargin=*, nosep, topsep=2pt, label=\arabic*.]
    \item We propose \textbf{WaveSpec}, a multispectral iris
    PAD framework that performs feature level spectral reasoning with a
    Set Conditioned Spectral Coherence Adapter. The adapter updates each
    band feature using the set consensus, the deviation from that
    consensus, and wavelength identity, allowing the model to detect
    artefact induced spectral inconsistencies directly in representation
    space.

    \item We introduce a multi head spectral representation that separates
    adapted structural information, consensus driven spectral cues, and
    deviation based anomaly cues. These representations are combined
    with adaptive weighted band pooling and trained with complementary
    classification, contrastive, prototype, spectral ordering, and
    anomaly separation objectives.

    \item We evaluate WaveSpec on the MSIrPAD~\cite{ramachandra2025spectrairispad} multispectral iris PAD
    benchmark under a strict cross artefact protocol, trained with one
    attack type and tested on unseen attack types, against strong
    baselines spanning conventional deep iris PAD, adversarial
    augmentation, foundation model based PAD, and prior multispectral
    fusion including SpectraIrisPAD. The analysis covers ablation,
    operating point evaluation using BPCER at fixed APCER levels, and
    computational benchmarking, showing that explicit spectral coherence
    modelling improves generalisation over late fusion and prior
    multispectral PAD strategies at modest overhead over the DINOv2
    backbone.
\end{enumerate}
\section{Related Work}
\label{sec:related_work}
This section reviews iris PAD for single band near infrared imaging
and then the multispectral methods that are most relevant to this
work.
\subsection{Near infrared iris presentation attack detection}
\label{sec:rw_nir_pad}

Iris presentation attack detection has been widely studied for near
infrared iris recognition systems. Early methods used hand crafted
texture descriptors, image quality measures, photometric cues, or
geometric information to detect printed irides, replayed images,
artificial eyes, and textured contact lenses. These approaches are
efficient and often interpretable, but their performance is sensitive to
sensor characteristics, segmentation quality, image normalisation, and
the attack materials observed during training. This limitation becomes
clear when PAD systems are evaluated on unseen presentation attack
instruments, as shown in the LivDet Iris evaluations~\cite{tinsley2023iris}.

Deep learning has improved iris PAD by learning features directly from
data. CNN based models such as DNetPAD, contact lens detection
networks, and adversarial augmentation strategies have shown strong
results under controlled settings. Recent work has also studied
cross-domain PAD, where the training and test data differ in sensor,
dataset, or attack type. ADV-GEN, for example, uses parametrically
generated adversarial samples to improve cross-domain iris PAD
performance~\cite{pal2025advgen}. Transformer based PAD models and
cascaded ViT designs have also been explored for stronger representation
learning~\cite{sharma2025cascading}. Despite these advances, most NIR
PAD methods still operate on a single spectral observation and therefore
cannot exploit wavelength dependent differences between bona fide tissue
and attack materials.

Visual foundation models have recently been considered for iris PAD.
DINOv2 and OpenCLIP features have been evaluated for NIR iris PAD with
lightweight task heads~\cite{tapia2025foundationiris}, while broader
biometric studies have tested vision language and multimodal foundation
models for iris recognition and iris PAD~\cite{sony2025benchmarking}.
Open-set studies such as VISER further show that unseen attack types
remain difficult even when strong pretrained representations or human
saliency cues are used~\cite{dowling2026viser}. These works support the
use of foundation features, but they do not explicitly model the
spectral relationships among multiple NIR bands.

\subsection{Multispectral iris PAD}
\label{sec:rw_multispectral_pad}

Multispectral iris PAD is motivated by the fact that bona fide irides
and presentation attack instruments do not respond to illumination
wavelength in the same way, which makes the spectral response across
NIR bands a useful cue when the attack type at test time differs from
those seen during training.

Earlier multispectral iris PAD methods used Purkinje reflections under
two wavelengths, gradient-based fusion across multiple NIR bands, or
reflectance cues near the iris--sclera boundary to detect artificial
eyes, print attacks, and textured contact lenses~\cite{purkinje_Image,
MS_Iris_PAD,MS_Iris_PAD_contactLens}. These studies demonstrated that
spectral information is useful, but they often relied on hand-crafted
measurements, specific acquisition conditions, or accurate localisation.
Later work introduced a near-infrared multispectral iris sensor to study
material-dependent spectral responses for recognition and PAD
~\cite{raghavendraMSIrisSensor}, and subsequent methods used five NIR
bands from 800\,nm to 980\,nm with CNN features for detecting textured
contact lens attacks~\cite{ramachandra2024multi, ramachandra2025spectrairispad}. While these approaches showed that
spectral diversity improves PAD, their attack coverage was limited and
the fusion mechanism did not explicitly model how the bands of the same
sample agree or deviate from one another.

SpectraIrisPAD extended this direction by introducing the MSIrPAD
benchmark and by using DINOv2 features with spectral position encoding
and token fusion for multispectral iris PAD~\cite{ramachandra2025spectrairispad}.
This was an important step because it showed that foundation-model
features and wavelength-aware conditioning can improve cross-artefact
performance. However, SpectraIrisPAD primarily treats wavelength
information as a conditioning and fusion cue: it does not explicitly
model whether each band agrees with, or deviates from, the shared
spectral behaviour of the same sample before classification.

WaveSpec addresses this limitation by moving from wavelength-conditioned
fusion to feature-level spectral coherence adaptation
(Sec.~\ref{sec:method}). This paper does not introduce a new database;
it uses the MSIrPAD protocol to study a different question: whether
explicit spectral coherence modelling improves generalisation to unseen
iris presentation attack instruments.

\section{Proposed Method}
\label{sec:method}

\begin{figure*}[t]
    \centering
    \includegraphics[width=\textwidth]{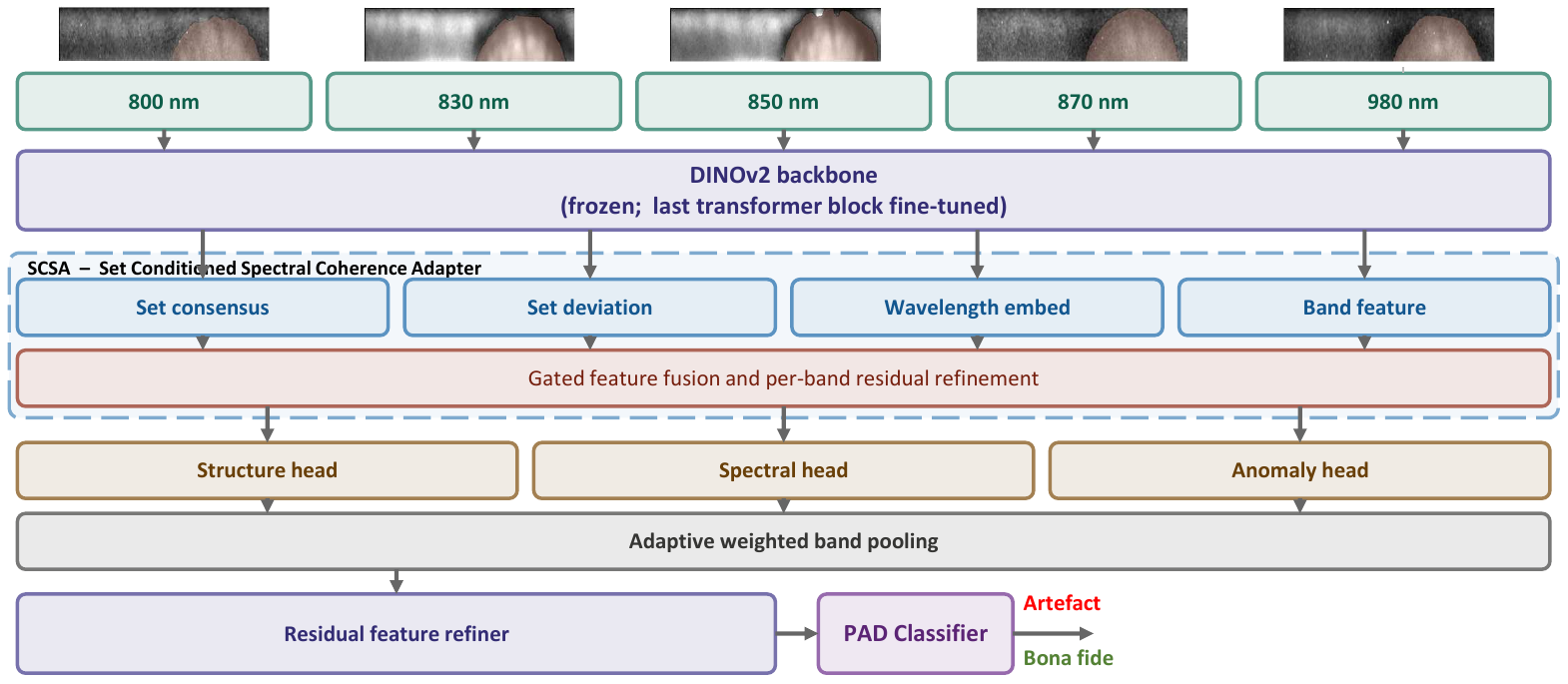}
    \caption{Overview of WaveSpec. Each available NIR band is encoded by
    DINOv2 and then adapted by the Set Conditioned Spectral Coherence
    Adapter. The adapter uses the consensus of the spectral set, the
    deviation of each band from that consensus, wavelength identity, and
    the original per band feature. The adapted representations are then
    projected, pooled, refined, and classified as bona fide or artefact.}
    \label{fig:Propo}
    \vspace{-2mm}
\end{figure*}
This section presents the motivation for WaveSpec, the per band
feature extraction, the proposed SCSA, and the pooling and decision
stages, followed by the training objective.
\subsection{Motivation and overview}
\label{sec:method:overview}

The motivation for WaveSpec is that an iris presentation is not a set of
independent NIR images but a spectral observation of the same physical
object. A bona fide iris is produced by biological tissue, whose
appearance across wavelengths is shaped by pigmentation, scattering, and
absorption, whereas an artefact is governed by pigments, inks,
substrates, or display surfaces that may look plausible at one
wavelength but respond inconsistently when viewed as a spectral set.
The useful PAD cue therefore lies not only in the appearance of a single
band, but in how the bands agree with, or depart from, one another.

WaveSpec is designed to model this relation before making the PAD
decision. Given up to $N{=}5$ NIR bands
$\nu \in \{800,830,850,870,980\}$\,nm, each available band is first
encoded independently with a mostly frozen DINOv2 backbone. The resulting
features are then passed to the proposed \emph{Set Conditioned Spectral
Coherence Adapter} (SCSA). For each band, SCSA builds an adapted feature
from four sources: the original DINOv2 feature, the consensus of the
available spectral set, the deviation of the band from that consensus,
and a learnable wavelength embedding. This design makes the classifier
sensitive to spectral coherence rather than only to per band appearance.
The adapted features are projected into structural, spectral, and
anomaly representations, pooled with learned band weights, and classified
as bona fide or attack.

Let $K$ be the number of available bands in a sample, with
$1 \leq K \leq N$. The DINOv2 feature dimension is $D{=}768$, and the
latent projection dimension is $L{=}256$. Since all set operations are
defined over the available bands, the same model can operate with a full
five band sample or with a partial spectral observation.

\subsection{Per band feature extraction}
\label{sec:method:backbone}

Each band image $I_k$ is a single channel NIR image. To use DINOv2, the
image is resized to $224{\times}224$ and replicated into three channels,
giving $\hat{I}_k \in \mathbb{R}^{3\times224\times224}$. We use
\textit{dinov2-base}~\cite{oquab2023dinov2} as the backbone.
All transformer blocks are frozen except the final block, which is fine-tuned jointly with the WaveSpec modules. This keeps the generic
representation learned by DINOv2 while allowing limited task adaptation,
and reduces overfitting risk on the comparatively small PAD dataset. The dinov2 base backbone is chosen to match
SpectraIrisPAD~\cite{ramachandra2025spectrairispad}, so that the
comparison in Sec.~\ref{sec:Exp} isolates the contribution of the set
level modelling rather than differences in backbone capacity; pairing WaveSpec with a smaller backbone is left for future work. 

For each band, the \texttt{[CLS]} token is used as the band feature:
\begin{equation}
    \mathbf{f}_k =
    \mathrm{DINOv2}(\hat{I}_k)[0] \in \mathbb{R}^{D},
    \qquad k=1,\ldots,K.
    \label{eq:dino}
\end{equation}
The features form the spectral set
$\mathbf{F}=\{\mathbf{f}_k\}_{k=1}^{K}$. The bands are deliberately
encoded independently at this stage. Cross wavelength reasoning is left
to SCSA, where it can be learned explicitly from the relation between
features.

\subsection{Set Conditioned Spectral Coherence Adapter}
\label{sec:method:scsa}

SCSA asks a simple question for each band: does this band agree with the
rest of the spectral observation? We first compute the set consensus
\begin{equation}
    \boldsymbol{\mu} =
    \frac{1}{K}\sum_{k=1}^{K}\mathbf{f}_k \in \mathbb{R}^{D},
    \label{eq:consensus}
\end{equation}
which represents the common information shared by the available bands.
We then compute the deviation of each band from this consensus:
\begin{equation}
    \mathbf{r}_k = \mathbf{f}_k-\boldsymbol{\mu}.
    \label{eq:deviation}
\end{equation}
The consensus captures what the spectral set agrees on, while
$\mathbf{r}_k$ captures how the $k$-th band differs from the shared
response. This residual is central to the method: for bona fide samples
the deviations are expected to be structured and moderate, whereas
artefacts can introduce irregular deviations due to material dependent
NIR behaviour.

Each wavelength also has a learnable embedding
$\mathbf{e}_k \in \mathbb{R}^{D}$. This gives the adapter knowledge of
which physical band produced the feature. The original feature,
consensus, deviation, and wavelength embedding are concatenated:
\begin{equation}
    \mathbf{x}_k =
    [\,\mathbf{f}_k \,\|\, \boldsymbol{\mu} \,\|\, \mathbf{r}_k \,\|\, \mathbf{e}_k\,]
    \in \mathbb{R}^{4D}.
    \label{eq:scsa_concat}
\end{equation}
Two lightweight MLPs then produce a spectral update and a gate:
\begin{equation}
    \boldsymbol{\delta}_k = \mathrm{MLP}_{\mathrm{proj}}(\mathbf{x}_k),
    \qquad
    \mathbf{g}_k =
    \sigma\!\left(\mathrm{MLP}_{\mathrm{gate}}(\mathbf{x}_k)\right).
    \label{eq:delta_gate}
\end{equation}
The adapted feature is obtained through a gated residual update,
\begin{equation}
    \tilde{\mathbf{f}}_k =
    \mathbf{f}_k + \mathbf{g}_k \odot \boldsymbol{\delta}_k .
    \label{eq:adapted}
\end{equation}
The gate is used because not every DINOv2 dimension should be changed.
Some dimensions already encode useful iris structure, while others
benefit from spectral context. The residual form therefore lets the
network introduce spectral information selectively instead of replacing
the backbone feature. The adapter also produces two specialist outputs:
a consensus representation $\mathbf{c}_k=\phi_c(\tilde{\mathbf{f}}_k)$
and an anomaly representation $\mathbf{a}_k=\phi_a(\mathbf{r}_k)$,
each through a LayerNorm and linear layer.

\subsection{Projection, pooling, and PAD decision}
\label{sec:method:heads_pool}

The adapted representation is decomposed into three complementary views.
The structure head projects $\tilde{\mathbf{f}}_k$ and preserves local
texture and shape cues. The spectral head projects the consensus
conditioned representation and is encouraged to respect wavelength
identity. The anomaly head projects the deviation signal and is intended
to respond to material induced inconsistencies. We write
\begin{equation}
\begin{aligned}
    \mathbf{s}_k &= \phi_s(\tilde{\mathbf{f}}_k),\\
    \boldsymbol{\rho}_k &= \phi_\rho(\mathbf{c}_k),\\
    \mathbf{z}_k &= \phi_z(\mathbf{a}_k),
\end{aligned}
\qquad
    \mathbf{h}_k =
    [\,\mathbf{s}_k \,\|\, \boldsymbol{\rho}_k \,\|\, \mathbf{z}_k\,]
    \in \mathbb{R}^{3L},
    \label{eq:heads}
\end{equation}
where each $\phi$ is a LayerNorm followed by a linear projection.

Different wavelengths may be more or less informative depending on the
attack material. For this reason, WaveSpec uses learned attention
pooling instead of simple averaging:
\begin{equation}
    w_k =
    \frac{\exp(\mathrm{MLP}_{\mathrm{pool}}(\mathbf{h}_k))}
    {\sum_{j=1}^{K}\exp(\mathrm{MLP}_{\mathrm{pool}}(\mathbf{h}_j))},
    \qquad
    \mathbf{p} = \sum_{k=1}^{K} w_k\mathbf{h}_k .
    \label{eq:pooling}
\end{equation}
The softmax normalisation makes the pooling valid for any number of
available bands. During training, we also apply band dropout after SCSA
with probability $0.20$, while ensuring that at least one band remains.
This prevents the classifier from relying on a fixed band configuration
and improves robustness to missing or unreliable spectral observations.

The pooled vector is refined with a residual MLP and passed to a
two class classifier:
\begin{equation}
    \tilde{\mathbf{p}} =
    \mathbf{p}+\mathrm{MLP}_{\mathrm{ref}}(\mathbf{p}), \qquad
    \hat{\mathbf{y}} = \mathrm{Classifier}(\tilde{\mathbf{p}}).
    \label{eq:classifier}
\end{equation}

\subsection{Training objective}
\label{sec:method:losses}

WaveSpec is trained with a compact multi term objective:
\begin{equation}
    \mathcal{L} =
    \mathcal{L}_{\mathrm{CE}}
    + \lambda_{\mathrm{SC}}\mathcal{L}_{\mathrm{SC}}
    + \lambda_{\mathrm{P}}\mathcal{L}_{\mathrm{P}}
    + \lambda_{\mathrm{SO}}\mathcal{L}_{\mathrm{SO}}
    + \lambda_{\mathrm{AS}}\mathcal{L}_{\mathrm{AS}},
    \label{eq:total_loss}
\end{equation}
with $\lambda_{\mathrm{SC}}{=}0.15$, $\lambda_{\mathrm{P}}{=}0.10$,
$\lambda_{\mathrm{SO}}{=}0.10$, and $\lambda_{\mathrm{AS}}{=}0.05$.
These weights were chosen from preliminary experiments during model
development, guided by keeping the cross entropy term dominant, and
then frozen for all reported experiments. No per artefact or per test
tuning is performed.
The cross entropy term is class balanced to handle the imbalance between
bona fide and attack samples. The supervised contrastive term encourages
same class samples to form compact regions in the refined feature space,
which helps when the test attack differs from the training attack. The
prototype loss adds global class anchors so that the representation does
not depend only on mini batch composition. The spectral ordering loss
encourages the spectral head to preserve wavelength identity and physical
proximity among bands. Finally, the anomaly separation loss encourages
small anomaly magnitude for bona fide samples and larger deviation
magnitude for attacks. Together, these losses shape the feature space so
that PAD decisions depend not only on class labels, but also on spectral
structure and material induced deviations.

\section{Implementation Details and Latency}
\label{sec:implem}

WaveSpec is implemented in PyTorch with the
\textit{dinov2-base} backbone at $224{\times}224$ input
resolution. All DINOv2 blocks are frozen except the final transformer
block. The trainable part of the model consists of this final DINOv2
block, SCSA, the projection heads, adaptive pooling, the residual
refiner, the classifier, and the learnable class and spectral
prototypes. This gives approximately $11.94$M trainable parameters:
about $7.09$M from the final DINOv2 block and $4.86$M from the
WaveSpec modules.

Training is performed for at most $20$ epochs using AdamW with learning
rate $3{\times}10^{-5}$, weight decay $10^{-4}$, and batch size $16$.
The learning rate is reduced by a factor of $0.2$ when validation
accuracy plateaus for two epochs, and early stopping is applied after
five epochs without improvement. We use light geometric augmentation,
including horizontal flipping, small rotation, mild affine jitter, and
set band dropout.

For a sample with $K$ available bands, inference requires $K$ DINOv2
forward passes followed by the lightweight SCSA and classification
heads; in the five band setting the total cost is approximately
$94.75$ GFLOPs per sample. Latency was measured in evaluation mode with
gradients disabled, using 30 warm up iterations and 300 timed CUDA event
runs at batch size $1$. On an NVIDIA L40S-48Q GPU, the five band
end-to-end latency is $11.15$ ms per sample ($89.6$ samples/s), of which
the DINOv2 backbone accounts for $9.17{\pm}0.34$ ms and SCSA with the
downstream heads for $1.57{\pm}0.07$ ms.

\begin{table*}[t]
\centering
\caption{Average cross-artefact PAD performance over the eight
train-on-one-artefact settings. Values are mean $\pm$ SD over training
artefacts. Lower is better.}
\label{tab:avg_cross_artifact}
\scriptsize
\setlength{\tabcolsep}{3.0pt}
\renewcommand{\arraystretch}{0.90}
\resizebox{\textwidth}{!}{%
\begin{tabular}{@{}lcccc@{}}
\toprule
Method & D-EER & BPCER@APCER=1\% & BPCER@APCER=5\% & BPCER@APCER=10\% \\
\midrule
CLIP~\cite{sony2025benchmarking} 
& $44.07{\pm}4.19$ 
& $97.65{\pm}2.89$ 
& $97.10{\pm}3.57$ 
& $95.71{\pm}5.31$ \\
DNetPAD~\cite{DNetPAD} 
& $33.66{\pm}6.68$ 
& $85.30{\pm}9.49$ 
& $74.81{\pm}12.73$ 
& $67.64{\pm}13.32$ \\
ViTPAD~\cite{sharma2025cascading} 
& $16.72{\pm}6.32$ 
& $52.72{\pm}13.79$ 
& $39.55{\pm}14.35$ 
& $30.83{\pm}13.42$ \\
ADV-GEN~\cite{pal2025advgen} 
& $18.99{\pm}8.76$ 
& $62.46{\pm}20.77$ 
& $46.05{\pm}22.50$ 
& $36.14{\pm}22.39$ \\
DeFu~\cite{ramachandra2024multi} 
& $14.51{\pm}6.74$ 
& $41.57{\pm}14.70$ 
& $30.56{\pm}13.39$ 
& $21.82{\pm}12.34$ \\
SpectraIrisPAD~\cite{ramachandra2025spectrairispad} 
& $8.45{\pm}4.28$ 
& $43.16{\pm}21.53$ 
& $31.44{\pm}21.83$ 
& $23.60{\pm}18.85$ \\
\textbf{WaveSpec} 
& $\mathbf{3.33{\pm}1.86}$ 
& $\mathbf{12.07{\pm}6.87}$ 
& $\mathbf{6.48{\pm}4.96}$ 
& $\mathbf{4.56{\pm}4.37}$ \\
\bottomrule
\end{tabular}%
}
\vspace{-1mm}
\end{table*}

\begin{figure}[t]
\centering
\includegraphics[width=0.99\columnwidth]{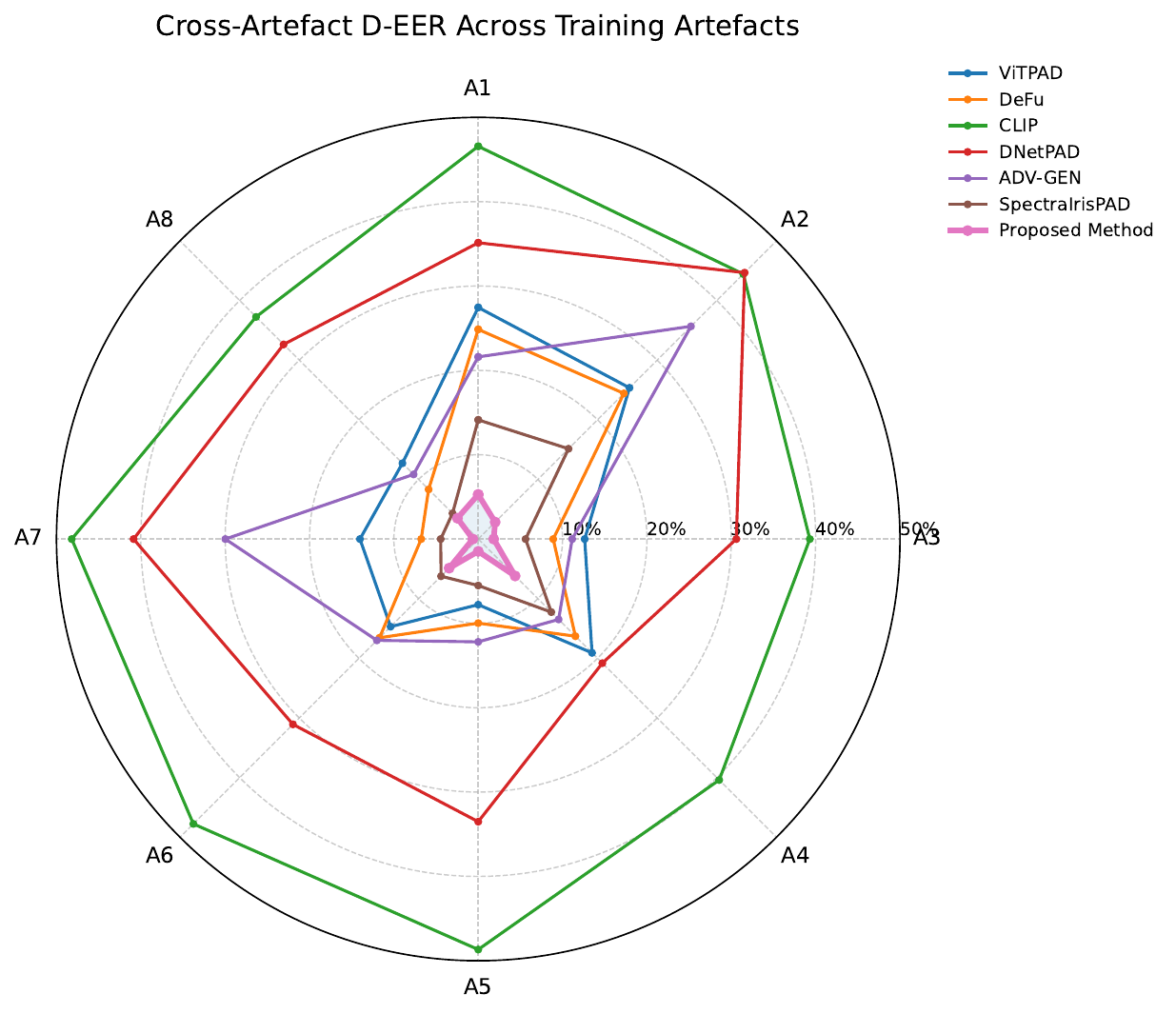}
\caption{Cross-artefact D-EER across the eight train-on-one-artefact
settings. Lower values are better, so curves closer to the centre
indicate stronger generalisation. WaveSpec remains consistently closer
to the centre than the competing PAD methods; the same trend holds at
the strict BPCER@APCER=1\% operating point (Table~\ref{tab:avg_cross_artifact}).
A1 to A8 denote the artefacts described in Sec.~\ref{sec:exp_protocol}.}
\label{fig:radar_cross_artifact}
\vspace{-2mm}
\end{figure}
\section{Experiments and Results}
\label{sec:Exp}
The experiments assess the main claim of this work: modelling spectral
coherence before classification improves generalisation to unseen iris
presentation attacks.  We first describe the MSIrPAD protocol and
evaluation metrics, then compare WaveSpec with existing PAD baselines,
and finally analyse the contribution of the main architectural
components through ablation.
\subsection{Dataset, protocol, and metrics}
\label{sec:exp_protocol}

We evaluate WaveSpec on the MSIrPAD benchmark introduced in
SpectraIrisPAD~\cite{ramachandra2025spectrairispad}. The database was
captured with a custom multispectral NIR iris sensor and contains five
spectral bands, namely 800, 830, 850, 870, and 980\,nm. The attack set
covers eight presentation attack instruments (PAIs): six textured
contact lens types (A1 to A6), one display attack (A7), and one print attack (A8). We use
the same identity-disjoint train, development, and test partitions as
SpectraIrisPAD~\cite{ramachandra2025spectrairispad} so that the comparison is directly aligned with the
previous benchmark.

The main evaluation follows a strict cross-artefact protocol. In each
experiment, the model is trained using bona fide samples and one attack
artefact, selected on the development set, and tested on all remaining
unseen artefacts. This setting is intentionally difficult because the
test attack material is not observed during training. It therefore
measures whether a PAD method can generalise to new presentation attack
instruments, rather than only recognising attacks already seen during
training.

We compare WaveSpec with CLIP~\cite{sony2025benchmarking},
DNetPAD~\cite{DNetPAD}, ViTPAD~\cite{sharma2025cascading},
ADV-GEN~\cite{pal2025advgen}, DeFu~\cite{ramachandra2024multi}, and
SpectraIrisPAD~\cite{ramachandra2025spectrairispad}. We report D-EER
and BPCER at APCER operating points of 1\%, 5\%, and 10\%~\cite{ISOPAD}, with the operating point thresholds calibrated on the development set and applied unchanged to the test data. Results are reported as mean performance over the eight train-on-one-artefact
settings, while the radar plot shows the corresponding per-artefact
trends. Lower values indicate better PAD performance. The full
per-training-artefact numerical table is provided in the supplementary
material.

\begin{table*}[t]
\centering
\caption{Cross-artefact ablation study of WaveSpec. Values are
mean $\pm$ SD in percent over the eight train-on-one-artefact settings.
For each training artefact, results are first averaged over the seven
unseen test artefacts. Lower is better. The full model row corresponds to the models reported in
Table~\ref{tab:avg_cross_artifact}. Retraining the identical
configuration with a different seed stream yields
$3.24{\pm}1.48$ D-EER, indicating stable results across seeds.}
\label{tab:ablation_cross_artifact}
\scriptsize
\setlength{\tabcolsep}{2.4pt}
\renewcommand{\arraystretch}{0.88}
\resizebox{\textwidth}{!}{%
\begin{tabular}{@{}lcccc@{}}
\toprule
Variant & D-EER & BPCER@APCER=1\% & BPCER@APCER=5\% & BPCER@APCER=10\% \\
\midrule
Full model & $3.33{\pm}1.86$ & $12.07{\pm}6.87$ & $6.48{\pm}4.96$ & $4.56{\pm}4.37$ \\
\midrule
w/o SCSA adapter        & $5.17{\pm}3.24$ & $16.48{\pm}9.81$  & $12.19{\pm}8.86$  & $9.05{\pm}7.00$  \\
w/o set consensus       & $5.97{\pm}4.56$ & $18.59{\pm}11.35$ & $13.34{\pm}12.78$ & $9.69{\pm}9.33$  \\
w/o spectral residual   & $4.36{\pm}3.05$ & $16.05{\pm}8.88$  & $9.63{\pm}8.47$   & $7.06{\pm}7.88$  \\
w/o wavelength embedding& $3.91{\pm}1.80$ & $12.65{\pm}3.75$  & $7.96{\pm}4.85$   & $6.17{\pm}4.94$  \\
w/o adaptive pooling    & $4.64{\pm}2.76$ & $17.83{\pm}14.52$ & $10.35{\pm}8.44$  & $7.87{\pm}7.14$  \\
w/o spectral ordering loss & $3.76{\pm}2.53$ & $10.98{\pm}7.87$ & $6.67{\pm}5.40$ & $5.53{\pm}4.96$ \\
w/o SupCon loss         & $4.00{\pm}2.07$ & $13.08{\pm}5.89$  & $8.59{\pm}5.15$   & $6.35{\pm}4.92$  \\
w/o prototype loss      & $4.39{\pm}3.07$ & $14.06{\pm}8.66$  & $8.90{\pm}7.34$   & $6.41{\pm}5.73$  \\
w/o anomaly loss        & $4.91{\pm}2.92$ & $15.92{\pm}11.39$ & $11.31{\pm}8.09$  & $9.38{\pm}7.40$  \\
w/o set-band dropout    & $3.94{\pm}2.37$ & $15.24{\pm}9.85$  & $9.02{\pm}7.31$   & $5.76{\pm}5.35$  \\
w/o all aux.\ losses    & $8.82{\pm}1.01$ & $17.45{\pm}3.19$  & $14.91{\pm}2.79$  & $11.62{\pm}2.50$ \\
\midrule
Consensus as median     & $4.36{\pm}1.56$ & $11.05{\pm}4.53$  & $7.92{\pm}4.22$   & $5.48{\pm}3.86$  \\
Consensus as maximum    & $6.23{\pm}4.58$ & $20.26{\pm}10.58$ & $14.75{\pm}10.03$ & $11.70{\pm}10.24$ \\
\bottomrule
\end{tabular}%
}
\vspace{-4mm}
\end{table*}
\subsection{Cross-artefact performance}
\label{sec:exp_cross_artifact}

Table~\ref{tab:avg_cross_artifact} summarises the average cross-artefact
performance, while Fig.~\ref{fig:radar_cross_artifact} shows how the
methods behave across the eight training artefacts. WaveSpec achieves
the lowest average error for every metric. It reduces mean D-EER from
8.45\% for SpectraIrisPAD to 3.33\%, corresponding to a 60.6\% relative
reduction. The improvement is even stronger at fixed operating points:
average BPCER is reduced from 43.16\% to 12.07\% at APCER=1\%, from
31.44\% to 6.48\% at APCER=5\%, and from 23.60\% to 4.56\% at
APCER=10\%. These reductions are important because low-APCER operation
is the most relevant regime for PAD deployment, where attack acceptance
must be tightly controlled while keeping bona fide rejection low.

The radar plot shows that the gain is not driven by a single favourable training artefact: the WaveSpec curve remains close to the centre across contact lens, display, and print attack settings, with the same trend at BPCER@APCER=1\% (Table~\ref{tab:avg_cross_artifact}). This indicates stable generalisation when the attack material changes. By contrast, CLIP and DNetPAD show high error for most artefacts, while DeFu, ViTPAD, ADV-GEN, and SpectraIrisPAD exhibit larger variations across the eight training conditions. Matched late score fusion of independently trained per-band DINOv2 models gives 17.60\% D-EER (Suppl. Table~2), compared with 8.45\% for SpectraIrisPAD and 3.33\% for WaveSpec, showing that score aggregation alone does not explain the gain. The proposed feature-level spectral coherence modelling instead compares each band with the consensus of the available spectral set before classification.

\subsection{Ablation study}
\label{sec:exp_ablation}

Table~\ref{tab:ablation_cross_artifact} analyses the contribution of
the main WaveSpec components over all eight train-on-one-artefact
settings. For each training artefact, results are first averaged over
the seven unseen test artefacts, and the table reports mean $\pm$ SD
across the eight training settings. Removing the SCSA adapter
increases D-EER from 3.33\% to 5.17\%, confirming that the
improvement is not due to the DINOv2 backbone alone. Removing the
set consensus, spectral residual, wavelength embedding, or adaptive
pooling also degrades performance across all four metrics, showing
that WaveSpec benefits from combining a shared spectral reference with
band-specific deviation and adaptive band aggregation.

The auxiliary losses exhibit a more differentiated behaviour. Removing individual
losses degrades performance modestly, with D-EER rising to between
3.76\% (w/o spectral ordering loss) and 4.91\% (w/o anomaly loss).
The spectral ordering loss is a partial exception: its removal raises
D-EER by 0.43 percentage points but yields slightly lower
BPCER@APCER=1\% (10.98\% vs 12.07\%), suggesting that it primarily
shapes the global decision boundary rather than the score distribution
at the tightest operating point. The full model is superior on the
remaining three metrics, confirming the loss provides consistent
benefit. Removing all auxiliary losses simultaneously raises D-EER to
8.82\% demonstrating that their
collective regularisation is not redundant with the CE objective and
cannot be compensated by the architectural components alone.

The ablation shows that the architectural components and the training
objectives contribute in a complementary way. Explicit spectral
consensus, band deviation, wavelength identity, and adaptive pooling
each support unseen artefact generalisation, and the auxiliary losses
are needed to shape the feature space in which these components
operate, as removing all of them raises the error the most. The
generalisation ability of WaveSpec therefore comes from the combination
of the set level architecture and the objectives that train it, not
from either part alone.

The last two rows of Table~\ref{tab:ablation_cross_artifact} examine
the choice of the consensus operator in Eq.~\ref{eq:consensus}. The
arithmetic mean is used because it is free of parameters, invariant to
the order of the bands, and defined for any number of available bands,
which the missing band setting requires. Replacing the mean with the
element wise median gives similar performance, which shows that the
method does not depend on the exact form of the central tendency
operator. Replacing it with the element wise maximum degrades all four
metrics, which shows that the consensus must represent the common
behaviour of the spectral set rather than its extreme values.

\subsection{Robustness to Band Subset and Missing-Band Analysis}
\label{sec:exp_band_subset}
\begin{table}[t]
\centering
\caption{Band subset and missing-band analysis for WaveSpec. The model
is trained with five bands and evaluated using restricted spectral
subsets, including all single-missing-wavelength four-band cases.
SpectraIrisPAD is shown only as the standard five-band reference. Values
are mean $\pm$ SD over the eight train-on-one-artefact settings. Lower
is better. The single-band DINOv2 baseline uses the same backbone, fine-tuning
depth, optimiser, augmentation, and applicable losses as WaveSpec, but
is trained and evaluated on one band.%
}
\label{tab:band_subset}
\scriptsize
\setlength{\tabcolsep}{3.0pt}
\renewcommand{\arraystretch}{0.90}
\resizebox{\columnwidth}{!}{%
\begin{tabular}{lcc}
\toprule
Setting & D-EER \% & BPCER@APCER=1\% \\
\midrule
SpectraIrisPAD, five bands & $8.45{\pm}4.28$ & $43.16{\pm}21.53$ \\
\midrule
Single-band DINOv2 baseline (980\,nm) & $22.53{\pm}0.96$ & $27.77{\pm}3.32$ \\
\midrule
WaveSpec, one band & $4.75{\pm}2.53$ & $14.44{\pm}5.88$ \\
WaveSpec, two bands & $3.92{\pm}1.94$ & $13.35{\pm}5.99$ \\
WaveSpec, three bands & $3.73{\pm}1.93$ & $13.10{\pm}6.56$ \\
WaveSpec, four bands & $3.53{\pm}1.91$ & $12.57{\pm}6.75$ \\
WaveSpec, five bands & $\mathbf{3.33{\pm}1.86}$ & $\mathbf{12.07{\pm}6.87}$ \\
\bottomrule
\end{tabular}%
}
\vspace{-8mm}
\end{table}

Table~\ref{tab:band_subset} evaluates whether WaveSpec can operate as
a spectral set model when only part of the wavelength set is available
at test time. Even with a single band, WaveSpec outperforms the
five band SpectraIrisPAD reference, and performance improves
monotonically as wavelengths are added, with the best result obtained
using all five bands. The single band baseline in
Table~\ref{tab:band_subset} shows that this behaviour is not explained
by the backbone and the training recipe alone. A DINOv2 model trained
with the same optimiser, augmentation, and applicable losses on a
single band reaches 22.53\% D-EER for the best performing band,
whereas WaveSpec evaluated with one
band reaches 4.75\%. The difference is that WaveSpec is trained on
multispectral sets, so the representation carries spectral information
learned through the set level components even when only one band is
observed at test time.

These results support two observations. First, WaveSpec does not depend
on a fixed five-band input and remains effective under partial spectral
observations, which follows from the SCSA design: consensus, deviation,
and adaptive pooling are computed over the available bands rather than
over a fixed image stack. Second, the consistent improvement from one to
five bands shows that the additional wavelengths provide complementary
cross-wavelength evidence for spectral coherence modelling.
\begin{figure}[t]
\centering
\includegraphics[width=0.99\columnwidth]{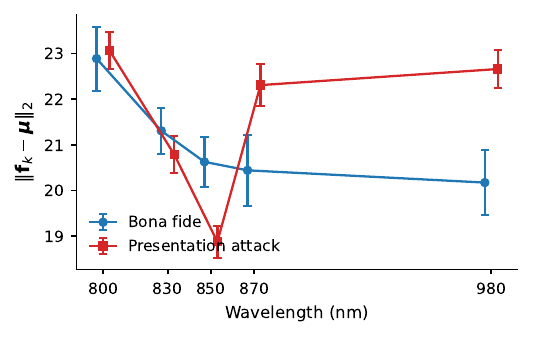}
\caption{Spectral residual profile over all complete five band test
captures (90 bona fide and 404 attack captures, the latter dominated by
display and print instruments). The residual of each band from the set
consensus is computed from frozen pretrained DINOv2 features without
any task training; error bars show the 95\% confidence interval of the
mean. The class difference is significant at 850, 870, and 980\,nm and
not at 800 and 830\,nm. Bona fide residuals decrease smoothly with
wavelength, whereas attack residuals are lowest at 850\,nm, the
standard iris imaging band, and rise sharply beyond it.}
\label{fig:residual_profile}
\vspace{-2mm}
\end{figure}

\subsection{Discussion and Limitations}
The experimental results suggest that the main benefit of WaveSpec comes
from modelling the spectral observation as a set, rather than as a group
of independent images. Figure~\ref{fig:residual_profile} supports this
view with a direct measurement of the coherence assumption. On complete
five band test captures, bona fide samples deviate less from their
spectral consensus than attacks on average, with a normalised deviation
of 0.410 against 0.434, and the two classes differ in the shape of the
deviation across wavelengths. Bona fide residuals decrease smoothly
from 800 to 980\,nm, while attack residuals reach their minimum at
850\,nm and rise sharply beyond it. This is consistent with attack
instruments being produced from, and optimised for, the standard iris
imaging band, and it indicates that the discriminative information lies
in the pattern of the deviation across bands rather than in its
magnitude alone, which is what the SCSA operates on. The distributions
of individual samples overlap substantially, so the residual magnitude
of a single band is not a classifier on its own, which is why the
deviation pattern is learned rather than thresholded.

The results also highlight the difficulty of multispectral iris PAD.
Although WaveSpec achieves the lowest average error, performance still
varies across training artefacts.  For WaveSpec, artefact~\#4 remains the most difficult
training condition, with higher variability than other artefacts (see Fig.~\ref{fig:radar_cross_artifact}).  This
suggests that some attack materials produce spectral responses closer to
bona fide iris tissue, or that the artefact-specific training data do not
fully cover the variability needed to separate all unseen attacks.  Therefore, the
important practical observation is: multispectral information improves
generalisation, but it does not remove the need for diverse attack
coverage during training.

A limitation of this work is that the evaluation is limited to MSIrPAD~\cite{ramachandra2025spectrairispad},
which is currently the only available multispectral iris PAD benchmark
with aligned five-band NIR images and diverse attack instruments. This
makes MSIrPAD suitable for studying spectral coherence, but it also
limits direct cross-dataset and cross-sensor validation in the same
multispectral setting. We therefore use the strict train-on-one,
test-on-unseen-artefacts protocol as the main measure of generalisation.
Future work should extend the evaluation as new multispectral iris PAD
datasets and sensors become available.

Finally, the contribution of this paper is the SCSA mechanism and its
analysis under cross-artefact, operating-point, ablation, latency, and
missing-band settings; the MSIrPAD dataset and protocol are inherited
from SpectraIrisPAD~\cite{ramachandra2025spectrairispad}. Future work
will study  uncertainty-aware
rejection under poor-quality bands, and domain adaptation for sensors
with different wavelength configurations.

\section{Conclusion}
\label{sec:Conc}
This paper presented WaveSpec, a multispectral iris PAD method that
models spectral coherence before classification. Unlike approaches that
mainly treat wavelength information as a conditioning or fusion cue,
WaveSpec adapts each band feature using the consensus of the available
spectral set, the deviation of that band from the consensus, and
wavelength identity. This enables the model to reason about whether a
multispectral observation behaves coherently as a bona fide iris or
shows material-induced inconsistencies associated with presentation
attack instruments.

Experiments on the MSIrPAD cross-artefact protocol show that WaveSpec
generalises better to unseen contact lens, display, and print attacks
than all compared state-of-the-art PAD methods, reducing mean D-EER from
8.45\% for the strongest prior multispectral baseline, SpectraIrisPAD,
to 3.33\%, with consistent gains across training artefacts and at all
reported operating points. The ablation and band-subset analyses confirm
that the gain stems from the explicit modelling of spectral consensus
and band-specific deviation rather than from the DINOv2 backbone alone,
and that WaveSpec remains effective under partial spectral observations.

{
    \small
    \bibliographystyle{ieeenat_fullname}
    \bibliography{Iris_PAD_Bib}
}

\end{document}